# Enhancing User Creativity: Semantic Measures for Idea Generation

Georgi V. Georgiev[1] and Danko D. Georgiev[2✉]

[1] Center for Ubiquitous Computing

Faculty of Information Technology and Electrical Engineering

University of Oulu, Oulu, Finland

E-mail: georgi.georgiev@oulu.fi

[2] Institute for Advanced Study, Varna, Bulgaria

E-mail: danko.georgiev@mail.bg

✉ Corresponding author:

Danko D. Georgiev, MD, PhD

Institute for Advanced Study

P.O. Box 504

Varna 9000, Bulgaria

E-mail: danko.georgiev@mail.bg

Tel: +359 877 095 888

**Highlights**

- Semantic networks could be used to quantify convergence and divergence in design thinking

- Successful ideas exhibit divergence of semantic similarity and increased information content in time

- Client feedback enhances information content and divergence of successful ideas

- Information content and semantic similarity could be monitored for enhancement of user creativity

**Abstract**

Human creativity generates novel ideas to solve real-world problems. This thereby grants us the power to transform the surrounding world and extend our human attributes beyond what is currently possible. Creative ideas are not just new and unexpected, but are also successful in providing solutions that are useful, efficient and valuable. Thus, creativity optimizes the use of available resources and increases wealth. The origin of human creativity, however, is poorly understood, and semantic measures that could predict the success of generated ideas are currently unknown. Here, we analyze a dataset of design problem-solving conversations in real-world settings by using 49 semantic measures based on WordNet 3.1 and demonstrate that a divergence of semantic similarity, an increased information content, and a decreased polysemy predict the success of generated ideas. The first feedback from clients also enhances information content and leads to a divergence of successful ideas in creative problem solving. These results advance cognitive science by identifying real-world processes in human problem solving that are relevant to the success of produced solutions and provide tools for real-time monitoring of problem solving, student training and skill acquisition. A selected subset of information content (IC Sanchez–Batet) and semantic similarity (Lin/Sanchez–Batet) measures, which are both statistically powerful and computationally fast, could support the development of technologies for computer-assisted enhancements of human creativity or for the implementation of creativity in machines endowed with general artificial intelligence.

1. **Introduction**

Creativity is the intellectual ability to create, invent, and discover, which brings novel relations, entities, and/or unexpected solutions into existence [1]. Creative thinking involves cognition (the mental act of acquiring knowledge and understanding through thought, experience, and senses), production, and evaluation [2]. We first become aware of the problems with which we are confronted, then produce solutions to those problems, and finally evaluate how good our solutions are. Each act of creation involves all three processes—cognition, production, and evaluation [2]. According to J. P. Guilford, who first introduced the terms convergence and divergence in the context of creative thinking, productive thinking can be divided into convergent and divergent thinking; the former which can generate one correct answer, and the latter which goes off in different directions without producing a unique answer [2]. Although currently there is no general consensus on the definition of convergent and divergent thinking, modern theories of creativity tend to have the following perspectives. Convergent thinking is regarded as analytical and conducive to disregarding causal relationships between items already thought to be related, whereas divergent thinking is viewed as associative and conducive to unearthing similarities or correlations between items that were not thought to be related previously [3-5].

Both convergent and divergent thinking are used to model the structure of intellect [6]. With regard to the nature of intelligence and originality, two general problem-solving behaviors were identified, those of the converger and those of the diverger, who exhibit convergent and divergent styles of reasoning/thinking, respectively [7]. The distinction between convergent and divergent thinkers is done based on the dimensions of scoring high on closed-ended intelligence tests versus scoring high on open-ended tests of word meanings or object uses [7]. The converger/diverger distinction also applies in cognitive styles and learning strategies [8]. Dual-processing accounts of human thinking see convergent and divergent styles as reflective/analytic and reflexive/intuitive, respectively [9], which is in line with current theories of creative cognition involving generation and exploration phases [10]. The convergent thinking style is assumed to induce a systematic, focused processing mode, whereas divergent thinking is suspected to induce a holistic, flexible task processing mode [11].

Psychological accounts that consider convergent and divergent production as separate and independent dimensions of human cognitive ability allow one to think of creative problem solvers as divergers rather than convergers [12], and to associate creativity with divergent thought that combines distant concepts together [13]. Focusing only on either convergent or divergent thinking, however, may inhibit the full understanding of creativity [14]. Viewing convergent production as a rational and logical process, and divergent production as an intuitive and imaginative process, creates the danger of oversimplification and confusion between intelligence and creativity. Instead, it should be recognized that there are parallel aspects or lines of thought that come together toward the end of the design process, making the design a matter of integration [14]. Since convergent and divergent thinking frequently occur together in a total act of problem solving [2], creativity may demand not only divergent thinking, but also convergent thinking [15, 16]. For example, deliberate techniques to activate human imagination rely on the elimination of criticism in favor of the divergent generation of a higher number of ideas. The process of deferred judgment in problem solving defers the evaluation of ideas and options until a maximum number of ideas are produced, thereby separating divergent thinking from subsequent convergent thinking [17]. This sequence of divergent and convergent thinking is classified as ideation-evaluation, where ideation refers to nonjudgmental imaginative thinking and evaluation to an application of judgement to the generated options during ideation [17]. Such accounts of creativity treat divergence and convergence as subsequent and iterated processes [18], particularly in that order. More recent accounts of creativity, however, highlight the interwoven role of both convergent and divergent thinking [15, 19, 20]. This interweaving has been identified in two ways. The analytic approach to creative problem solving based on linkography showed that convergent and divergent thinking are so frequent at the cognitive scale that they occur concurrently in the ideation phase of creative design [15]. The computational approach demonstrated that a computer program (comRAT-C), which uses consecutive divergence and convergence, generates results on a common creativity test comparable to the results obtained with humans [20]. Hence, the creative problem solver or designer may need to learn, articulate, and use both convergent and divergent skills in equal proportions [14].

The concurrent occurrence of convergent and divergent thinking in creative problem solving raises several important questions. Is it possible to evaluate convergence and divergence in problem-solving conversations in an objective manner? How do convergence and divergence relate to different participants in a problem-solving activity? Are there particular moments in the process of real-world problem solving where a definitive change from convergence to divergence, or vice versa, occurs? How do convergence and divergence relate to the success of different ideas that are generated and developed in the process of problem solving? Could semantic measures predict the future success of generated ideas, and can they be reverse-engineered to steer generated ideas toward success in technological applications, such as in computer-assisted enhancements of human creativity or implementations of creativity in machines endowed with artificial intelligence?

We hypothesized that semantic measures can be used to evaluate convergence and divergence in creative thinking, changes in convergence/divergence can be detected in regard to different features of the problem-solving process, including participant roles, successfulness of ideas, first feedback from client, or first evaluation by client or instructor, and semantic measures can be identified whose dynamics reliably predicts the success or failure of generated ideas. To test our hypotheses we analyzed the transcripts of design review conversations recorded in real-world educational settings at Purdue University, West Lafayette, Indiana, in 2013 [21]. The conversations between design students, instructors, and real clients, with regard to a given design task, consisted of up to 5 sessions (Table 1) that included the generation of ideas by the student, external feedback from the client, first evaluation by the client or instructor, and evaluation of the ideas by the client. The problem-solving conversations were analyzed in terms of participant role, successfulness of ideas, first feedback from client, or first evaluation by client or instructor using the average values of 49 semantic measures quantifying the level of abstraction (1 measure), polysemy (1 measure) or information content (7 measures) of each noun, or the semantic similarity (40 measures) between any two nouns in the constructed semantic networks based on WordNet 3.1 [22].

## 2. Materials and Methods

### 2.1. Design review conversations

Real-world conversations are an outstanding source to gain insights into the constructs of problem solving and decision making. To study human reasoning and problem solving, we focused on design review conversation sessions in real-world educational settings. The conversation sessions were between students and experienced instructors, and each session was used to teach and assess the student's reasoning and problem solving with regard to a given design task for a real client. The experimental dataset of design review conversations employed in this study was provided as a part of the 10th Design Thinking Research Symposium [21]. Here, we analyzed two subsets, with participants (students) majoring in Industrial Design: Junior (J): 1 instructor, 7 students (indicated with J1 to J7), and 10 other stakeholders (4 clients and 6 experts) and Graduate (G): 1 instructor, 6 students (indicated with G1 to G6), and 6 other stakeholders (2 clients and 4 other students).

The experimental dataset included data collected either from the same students and teams over time (although not always possible) or from multiple students and teams [21]. In addition, efforts were made to be gender inclusive. All data were collected *in situ* in natural environments rather than controlled environments. In some cases, the design reviews were conducted in environments well insulated from disruptive noises, surrounding activities, and lighting changes; in other cases, these conditions were not possible to achieve. When disruptions occurred, most were less than a minute in duration. Because English was a second language for a number of the participants, there were some light accents in the digital recordings [21]. The purpose of the conversations was for the instructor to notice both promising and problematic aspects in the student work and to help the students deal with possible challenges encountered [21]. At the end of these conversations, the students developed a solution (design concept for a product or service) that answered the problem posed in the task given initially.

Computational quantification of the results was based on the digital recordings and the corresponding written transcripts of the conversations. Because our main focus was on studying ideas in creative problem solving, we had explicitly defined the term *idea* as a formulated creative solution (product concept) to the given design problem (including product name, drawings of the product, principle of action, target group, etc.) [23, 24]. As an example, on the graduate project "Outside the Laundry Room," some of the generated ideas were "Laundry Rocker," "Clothes Cube," "Drying Rack," "Tree Breeze," "Washer Bicycle," etc. Our criterion for a minimal conversation was a conversation containing at least 15 nouns. Since on average 13.4% of the words in the conversation were nouns, an average minimal conversation contained ≈ 110 words. The reported results were per student and solution (idea).

### 2.1.1. Comparison between student thinking and instructor thinking

On the basis of the participant roles, the speech in the conversations was divided into speech by students or speech by instructors. Instructors were defined as those giving feedback or critique that were not only persons directly appointed as instructors in the particular setting, but also clients, sometimes other students acting or criticizing as instructors, or other stakeholders present on the intermediate or the final meetings. If there were several instructors in a conversation, their speech was taken together. For this comparison, the J and G subsets contained 7 and 6 subject cases, respectively, for a total of 13 cases. For both students and instructors, 39 conversation transcripts were each analyzed (Table 1).

*2.1.2. Comparison between successful ideas and unsuccessful ideas*

Conversations were divided into 2 groups: those related to unsuccessful ideas and those related to successful ideas. The unsuccessful were ideas that had not been developed to the end or had been disregarded in the problem-solving process, whereas successful ideas were those that had been developed to the end. The final evaluation of successful ideas was performed by the clients. For each student, only one of the generated ideas was the successful one. The same conversation was divided into a part or parts that concerned one or more unsuccessful ideas, and a part that concerned the successful idea. These divisions were made on sentence breaks. When two ideas were compared in one sentence, the sentence was considered to belong to the idea that was main for the comparison. In rare cases, if the main idea could not be identified, the sentence was not included in the analysis. The division of the text in the conversation transcripts between different ideas was assisted by the available slides in the dataset containing drawings of the generated ideas (product concepts), product names, principle of action, etc. For this comparison, the J and G subsets contained 7 and 5 subject cases, respectively, for a total of 12 cases. One case in the G subset was omitted because of missing data (slides with design sketches for client review) pertaining to unsuccessful ideas. For the 12 subject cases, the J subset contained conversations pertaining to 22 unsuccessful and 7 successful ideas; the G subset contained conversations pertaining to 19 unsuccessful and 5 successful ideas. In total, conversations pertaining to 41 unsuccessful ideas and 12 successful ideas were analyzed (Table 1).

*2.1.3. Comparison of ideas before and after first feedback*

Conversations were divided into 2 groups: containing ideas before and after first feedback. The division was based on a predefined point, which was the first feedback from the client (a stakeholder that was not a student or appointed as an instructor). For this comparison, the J and G subsets contained 7 and 5 subject cases, respectively, for a total of 12 cases. One case in the G subset was omitted due to missing data for ideas after the first feedback. For the 12 subject cases, the before first feedback group contained 25 conversation transcripts, whereas the after first feedback group contained 24 conversation transcripts (Table 1). The effect of first feedback on the time dynamics of successful ideas was assessed on 7 successful ideas (G4, G5, G6, J3, J5, J6, and J7) that had sufficiently long conversations to allow for the division into 6 time points comprising 2 sets of 3 time points before and after the first feedback.

### 2.1.4. Comparison of ideas before and after first evaluation

Conversations were divided into 2 groups: containing ideas before and after the first evaluation. The division was based on a predefined point, which was the first evaluation performed by the instructor (for the J subset) or the client (for the G subset). At the time of first evaluation, some of the generated ideas were discarded as unsuccessful. Those ideas that passed the first evaluation were developed further, mainly with a focus on details rather than on change of the main characteristics. In the G subset, two or more ideas passed the first evaluation, whereas in the J subset, only the successful idea passed the first evaluation. For this comparison, the J and G subsets contained 6 and 5 subject cases, respectively, for a total of 11 cases. One case of the J subset and one case of the G subset were omitted because of missing data for ideas after the first evaluation. For the 11 subject cases, the before first evaluation group contained 22 conversation transcripts, whereas the after first evaluation group contained 13 conversation transcripts (Table 1). The effect of first evaluation on the time dynamics of successful ideas was assessed on 8 successful ideas (G4, G5, G6, J1, J3, J4, J6, and J7) that had sufficiently long conversations to allow for the division into 6 time points comprising 2 sets of 3 time points before and after the first evaluation.

### 2.2. Modelling with semantic networks

In psychology, semantic networks depict human memory as an associative system wherein each concept can lead to many other relevant concepts [25]. In artificial intelligence, the semantic networks are computational structures that represent meaning in a simplified way within a certain region of conceptual space. The semantic networks consist of nodes and links. Each node stands for a specific concept, and each link, whereby one concept is accessed from another, indicates a type of semantic connection [25]. Semantic networks can be used to computationally model conceptual associations and structures [26, 27]. In this study, to construct semantic networks of nouns used in the conversations, we first cleaned the transcripts of the conversations for any indications of non-verbal expressions, such as "[Laughter]," speaker names and all the time stamps. As a second step, we processed the textual data using part-of-speech tagging performed by the Natural Language Toolkit (NLTK) [28] with the TextBlob library [29]. Then, we extracted only the nouns, both singular and plural. With the use of Python scripts, we processed all the nouns by converting the plural forms to singular and by removing nouns that were not listed in WordNet. In total, only 8 nouns were removed, which comprised 0.2% of all nouns that were analyzed. Finally, we analyzed the constructed semantic networks using WordNet 3.1.

### *2.3.   Analysis of time dynamics of semantic measures*

For graph analysis, we used Wolfram Mathematica, a mathematical symbolic computation program developed by Wolfram Research (Champaign, Illinois). The average level of abstraction, polysemy, information content and semantic similarity in the semantic network were computed using WordGraph 3.1, a toolset that implements the WordNet 3.1 is-a hierarchy of nouns as a directed acyclic graph, allowing for efficient computation of various graph-theoretic measures in Wolfram Mathematica. The is-a relationship between noun synsets (sets of synonyms) organizes WordNet 3.1 into a hierarchical structure wherein if synset A is a kind of synset B, then A is the hyponym of B, and B is the hypernym of A. As an example, the synset {cognition, knowledge, noesis} is a kind of {psychological_feature}.

The level of abstraction is negatively related to the depth of the noun in the taxonomy in a way that the root noun "entity" is the most abstract, whereas the deepest nouns in the taxonomy are least abstract [30]. The complement of the level of abstraction to unity is a measure of word concreteness.

The polysemy counts the number of meanings of each word, and its log-transformed value measures the bits of missing information that are needed by the listener to correctly understand the intended meaning of a given word.

The information content (IC) of nouns was calculated using seven IC formulas by Blanchard [31], Meng [32], Sanchez [33], Sanchez–Batet [34], Seco [35], Yuan [36], or Zhou [37].

The semantic similarity of pairs of nouns was calculated using five path-based similarity formulas by Al-Mubaid–Nguyen [38], Leacock–Chodorow [39], Li [40], Rada [41], or Wu–Palmer [42] and five IC-based similarity formulas by Jiang–Conrath [43], Lin [44], Meng [45], Resnik [46], or Zhou [47], each of which could be combined with any of the seven IC formulas, thereby generating 35 IC-based similarity measures. Because WordNet 3.1 as a database is much richer than a mathematical graph, we created and employed WordGraph 3.1, a custom toolset for Wolfram Mathematica that allows for fast and efficient computation of all graph-theoretic measures related to the is-a hierarchy of nouns.

To test whether convergent or divergent thinking could be quantified through convergence or divergence of semantic similarity, we assessed the change of the average semantic similarity in time. Convergence in the semantic networks was defined as an increase in the average semantic similarity in time (positive slope of the trend line), whereas divergence as a decrease in the average semantic similarity in time (negative slope of the trend line). To obtain 3 time points for analysis of time dynamics for each subject (Table 1), we joined the conversation transcripts pertaining to each group or idea and then divided the resulting conjoined conversations into 3 equal parts based on word count. This division was made into whole sentences in such a way that no time point of the conversation contained less than 5 nouns. Then, we assessed the time dynamics using linear trend lines. Because only nouns in the conversations were used for the construction of semantic networks, each time point had to contain at least 5 nouns to obtain a proper average semantic similarity.

### 2.4. *Semantic measures based on WordNet 3.1*

The calculation of semantic measures based on WordNet 3.1 (https://wordnet.princeton.edu/) was performed with the WordGraph 3.1 custom toolset for Wolfram Mathematica. The structure of WordGraph 3.1 is isomorphic to the is-a hierarchy of nouns in WordNet 3.1, implying that all mathematical expressions defined in WordGraph 3.1 also hold for WordNet 3.1. The nouns in WordGraph 3.1 were represented by 158441 case-sensitive word vertices (including spelling variations, abbreviations, acronyms, and loanwords from other languages) and 82192 meaning vertices, in which each word could have more than one meaning (polysemy) and each meaning could be expressed by more than one word (synset). WordGraph 3.1 consists of two subgraphs, subgraph $M$, which contains 84505 hypernym $\rightarrow$ hyponym edges between meaning vertices, and subgraph $W$, which contains 189555 word $\rightarrow$ meaning edges between word vertices and each of their meaning vertices.

Several graph-theoretic functions were used as follows:

Subvertices$(G, x)$: the subvertices of a vertex $x$ in a directed graph $G$ are all vertices in $G$ that have a finite directed path from $x$. Thus, every vertex is a subvertex of itself.

Subsumers$(G, x)$: the subsumers of a vertex $x$ in a directed graph $G$ are all vertices in $G$ that have a finite directed path to $x$. Thus, every vertex is a subsumer of itself.

Leaves$(G, x)$: a leaf in a directed graph $G$ is a vertex with a vertex out-degree of zero. In other words, the leaf does not have outgoing edges. The leaves of a vertex $x$ in a directed graph $G$ are all subvertices of $x$ with a vertex out-degree of zero. Because every vertex is a subvertex of itself, it follows that the number of leaves of each leaf in $G$ is 1.

ShortestPathDistance$(G, x, y)$: the shortest path distance between a vertex $x$ and a vertex $y$ in a directed graph $G$ is the minimal number of edges needed for a trip from $x$ to $y$. The shortest path distance is infinite $\infty$ if there is no path from $x$ to $y$. In general, the shortest path distance from $x$ to $y$ is not the same as the shortest path distance from $y$ to $x$; these distances are equal in undirected (bidirectional) graphs.

Depth$(G, x)$: the depth of a vertex $x$ in a rooted directed graph $G$ is 1 + the shortest path distance from the root vertex $r$ to $x$. Thus, the depth of the root vertex is 1.

VertexEccentricity$(G, x)$: the vertex eccentricity of a vertex $x$ in a directed graph $G$ is the length of the longest of all the shortest paths from the vertex $x$ to every other vertex in the graph $G$.

MaxDepth$(G)$: the maximal depth of a rooted directed graph $G$ is 1 + the vertex eccentricity of the root vertex $r$.

IncidenceList$(G, x)$: gives a list of all edges (incoming, outgoing, or undirected) incident to a vertex $x$ in a graph $G$.

With the use of the above graph-theoretic functions, semantic functions were constructed that take words as arguments and return values that depend only on the relationship between the word arguments and the meanings subgraph $M$ (Fig. 1). Two graph operators were used: $R(G)$

reverses the direction of all directed edges in the graph $G$, and $U(G)$ converts all directed edges in the graph $G$ into undirected (bidirectional) edges.

$|f(x)|$: gives the number of elements contained by the list $f(x)$.

$\text{Polysemy}(x) = |\text{IncidenceList}(W, x)|$: gives the number of all the meaning vertices that are 1 edge away from a given word $x$ (Fig. 1A).

$\text{Depth}(x)$: gives the shortest path distance between the root meaning vertex corresponding to the word "entity" and a word $x$ in the graph $M \cup \text{IncidenceList}[R(W), x]$ (Fig. 1A,C). Thus, the depth of the word "entity" is 1.

$\text{AbstractionLevel}(x)$: gives the level of abstraction of the word $x$ defined as $1 - \dfrac{\text{Depth}(x) - 1}{\text{Max\_depth} - 1}$.

$\text{Subsumers}(x)$: gives a list of the meaning subsumers of the word $x$ in the graph $M \cup \text{IncidenceList}[R(W), x]$, excluding $x$ itself since it is a word subsumer (Fig. 1A).

$\text{Subvertices}(x)$: gives a list of the meaning subvertices of the word $x$ in the graph $M \cup \text{IncidenceList}(W, x)$, excluding $x$ itself since it is a word subvertex (Fig. 1B).

$\text{Leaves}(x)$: gives a list of the leaves of the word $x$ in the graph $M \cup \text{IncidenceList}(W, x)$ (Fig. 1B).

$\text{Commonness}(x)$: the commonness of a word $x$ in the graph $G = M \cup \text{IncidenceList}(W, x)$ is defined as $\sum_{i \in \text{Leaves}(G, x)} \dfrac{1}{\text{Subsumers}(M, i)}$ (Fig. 1B).

$\text{LCS}(x, y)$: for $x \neq y$ gives the lowest common subsumer of a word $x$ and a word $y$ in the graph $G = M \cup \text{IncidenceList}[R(W), x] \cup \text{IncidenceList}[R(W), y]$ (Fig. 1C). The lowest common subsumer is a meaning vertex with maximal depth in the taxonomy among all vertices $z$ that minimize the sum $\text{ShortestPathDistance}[G, z, x] + \text{ShortestPathDistance}[G, z, y]$. If there is a tie between two or more common subsumers of $x$ and $y$, which are equally deep in the taxonomy, the uniqueness of $\text{LCS}(x, y)$ is ensured by taking the meaning vertex with the lowest entry number in WordNet 3.1.

$\text{Depth}[\text{LCS}(x, y)]$: gives the shortest path distance between the root word "entity" and a meaning vertex $\text{LCS}(x, y)$ in the graph $M \cup \text{IncidenceList}(W, \text{"entity"})$ (Fig. 1C).

$\text{Distance}(x, y)$: for $x \neq y$ gives the shortest path distance between a word $x$ and a word $y$ in the graph $U(M) \cup \text{IncidenceList}[U(W), x] \cup \text{IncidenceList}[U(W), y]$ minus 2 edges to subtract edge contribution outside of the meanings subgraph $M$ (Fig. 1D).

For the calculation of intrinsic information content of nouns, were used several constants that are specific for WordNet 3.1:

$\text{Max\_vertices}$: total number of meaning vertices is 82192.

$\text{Max\_leaves}$: total number of leaves is 65031.

$\text{Max\_depth}$: maximal depth of the taxonomy is 19.

$\text{Min\_commonness}$: minimal commonness of the word "Saint Ambrose" is 1/35.

$\text{Max\_commonness}$: maximal commonness of the root word "entity" is 6863.6.

### 2.4.1. Information Content (IC) measures

The intrinsic information content (IC) of a word $x$ in WordNet 3.1 was computed using seven different formulas:

IC by Blanchard et al. [31], normalized in the interval [0,1], is

$$IC(x) = 1 - \frac{\log|\text{Leaves}(x)|}{\log(\text{Max\_leaves})}$$

IC by Meng et al. [32]

$$IC(x) = \frac{\log[\text{Depth}(x)]}{\log(\text{Max\_depth})} \left[ 1 - \frac{\log\left[1 + \sum_{i \in \text{Subvertices}(x)} \frac{1}{\text{Depth}(i)}\right]}{\log(\text{Max\_vertices})} \right]$$

IC by Sanchez et al. [33], normalized in the interval [0,1], is

$$IC(x) = \frac{\log\left(\frac{|\text{Leaves}(x)|}{\text{Max\_leaves} \times |\text{Subsumers}(x)|}\right)}{\log\left(\frac{\text{Min\_commonness}}{\text{Max\_leaves}}\right)}$$

IC by Sanchez–Batet [34], normalized in the interval [0,1], is

$$IC(x) = \frac{\log\left[\frac{\text{Commonness}(x)}{\text{Max\_commonness}}\right]}{\log\left(\frac{\text{Min\_commonness}}{\text{Max\_commonness}}\right)}$$

IC by Seco et al. [35], normalized in the interval [0,1], is

$$IC(x) = 1 - \frac{\log|\text{Subvertices}(x)|}{\log(\text{Max\_vertices})}$$

IC by Yuan et al. [36]

$$IC(x) = \frac{\log[\text{Depth}(x)]}{\log(\text{Max\_depth})} \left(1 - \frac{\log|\text{Leaves}(x)|}{\log(\text{Max\_leaves})}\right) + \frac{\log|\text{Subsumers}(x)|}{\log(\text{Max\_vertices})}$$

IC by Zhou et al. [37]

$$IC(x) = \frac{1}{2}\left[1 - \frac{\log|\text{Subvertices}(x)|}{\log(\text{Max\_vertices})} + \frac{\log[\text{Depth}(x)]}{\log(\text{Max\_depth})}\right]$$

### 2.4.2. Path-based Similarity measures

The semantic similarity between a pair of words $x$ and $y$ such that $x \neq y$ was computed using five different path-based similarity formulas:

Al-Mubaid–Nguyen similarity [38], normalized in the interval [0,1], is

$$\text{sim}(x, y) = 1 - \frac{\log\left[1 + \text{Distance}(x, y) \times \left(\text{Max\_depth} - \text{Depth}\left[\text{LCS}(x, y)\right]\right)\right]}{\log\left[1 + 2\left(\text{Max\_depth} - 1\right)^2\right]}$$

Leacock–Chodorow similarity [39], normalized in the interval [0,1], is

$$\text{sim}(x, y) = 1 - \frac{\log\left[\text{Distance}(x, y) + 1\right]}{\log\left[2\left(\text{Max\_depth}\right) - 1\right]}$$

Li et al. similarity [40], normalized in the interval [0,1], is

$$\text{sim}(x, y) = e^{-0.2\,\text{Distance}(x, y)} \frac{e^{1.2\,\text{Depth}[\text{LCS}(x, y)]} - 1}{e^{1.2\,\text{Depth}[\text{LCS}(x, y)]} + 1}$$

Rada et al. similarity [41], normalized in the interval [0,1], is

$$\text{sim}(x, y) = 1 - \frac{\text{Distance}(x, y)}{2\left(\text{Max\_depth} - 1\right)}$$

Wu–Palmer similarity [42], normalized in the interval [0,1], is

$$\text{sim}(x, y) = \frac{2\left[\text{Depth}\left[\text{LCS}(x, y)\right] - 1\right]}{2\left[\text{Depth}\left[\text{LCS}(x, y)\right] - 1\right] + \text{Distance}(x, y)}$$

*2.4.3. IC-based Similarity measures*

The semantic similarity between a pair of words $x$ and $y$ such that $x \neq y$ was computed using five different IC-based similarity formulas, each of which was combined with every of the seven IC formulas thereby generating a total of 35 different IC-based similarity measures:

Jiang–Conrath similarity [43]

$$\text{sim}(x, y) = 1 - \frac{\left[IC(x) + IC(y) - 2IC\left[\text{LCS}(x, y)\right]\right]}{2}$$

Lin similarity [44]

$$\text{sim}(x, y) = \frac{2IC[\text{LCS}(x, y)]}{IC(x) + IC(y)}$$

Meng similarity [45]

$$\text{sim}(x, y) = \left[\frac{2IC[\text{LCS}(x, y)]}{IC(x) + IC(y)}\right]^{\frac{1-\exp[-0.08\,\text{Distance}(x, y)]}{\exp[-0.08\,\text{Distance}(x, y)]}}$$

Resnik similarity [46]

$$\text{sim}(x, y) = IC[\text{LCS}(x, y)]$$

Zhou similarity [47]

$$\text{sim}(x, y) = 1 - \frac{1}{2}\left[1 - \frac{\log[\text{Distance}(x, y) + 1]}{\log[2(\text{Max\_depth}) - 1]}\right] - \frac{1}{4}[IC(x) + IC(y) - 2IC[\text{LCS}(x, y)]]$$

### 2.5. *Statistics*

Statistical analyses of the constructed semantic networks were performed using SPSS ver. 23 (IBM Corporation, New York, USA). To reduce type I errors, the time dynamics of semantic measures were analyzed with only two a priori planned linear contrasts [48] for the idea type (sensitive to vertical shifts of the trend lines) or the interaction between idea type and time (sensitive to differences in the slopes of the trend lines). Because semantic similarity was calculated with 40 different formulas and information content with 7 different formulas, possible differences in semantic similarity or information content were analyzed with three-factor repeated-measures analysis of variance (rANOVA), where the idea type was set as a factor with 2 levels, the time was set as a factor with 3 levels, and the formula type was set as a factor with 40 or 7 levels, respectively. Differences in the average level of abstraction, polysemy, or individual measures of information content or semantic similarity were analyzed with two-factor rANOVAs, where the idea type and time were the two only factors. The implementation of the repeated-measures experimental design controlled for factors that cause variability between subjects, thereby simplifying the effects of the primary factors (ideas and time) and enhancing the power of the performed statistical tests. Pearson correlation analyses and hierarchical clustering of semantic similarity and IC measures were performed in R ver. 3.3.2 (R Foundation for Statistical Computing, Vienna, Austria). For all tests, the significance threshold was set at 0.05.

## 3. Results

### 3.1. *Student and instructor thinking are similar in terms of semantic measures*

With regard to creative thinking, our primary interest was focused on semantic similarity because as a two-argument function, it is able to evaluate the relationship between pairs of vertices in the constructed semantic networks. In addition, the average of semantic similarity is more informative than is the average of single-argument functions, such as information content, polysemy, or level of abstraction, because there are $(n^2 - n)/2$ pairs of vertices versus only $n$ vertices in the semantic network. A comparison between the student and instructor speech in the problem-solving conversations did not show significant differences in semantic similarity (three-factor rANOVA: $F_{1,12}<0.3$, $P>0.58$; Fig. 2A), information content (three-factor rANOVA: $F_{1,12}<0.2$, $P>0.65$; Fig. 2B), polysemy ($F_{1,12}<0.6$, $P>0.46$; Fig. 2C), or level of abstraction ($F_{1,12}<0.9$, $P>0.38$; Fig. 2D); this could be because all of the ideas originating from the student or the instructor were commented upon by both participants. To reduce Type II errors, we also confirmed that the linear contrasts in individual two-factor rANOVAs were not significant for each of the 40 semantic similarity measures ($F_{1,12}<0.9$, $P>0.37$) and each of the 7 information content measures ($F_{1,12}<0.8$, $P>0.40$). These results justify our decision to further analyze both student and instructor speech jointly with regard to different types of ideas contained in the conversations.

### 3.2. *Divergence of semantic similarity predicts the success of ideas*

Creative ideas should be novel, unexpected, or surprising, and provide solutions that are useful, efficient, and valuable [49, 50]. The success of generated ideas in creative problem solving depends not only on the final judgment by the client who decides which idea is the most creative, but also on the prior decisions made by the designer not to drop the idea in face of constraints on available physical resources. Thus, while success and creativity are not the same, the ultimate goal of design practice is to find solutions that are both creative and successful. To determine whether different types of thinking are responsible for the success of some of the generated ideas and the failure of others, we have compared the time dynamics of semantic measures in the conversations pertaining to successful or unsuccessful ideas. Three-factor rANOVA detected a significant crossover interaction between idea type and time ($F_{1,11}=11.4$, $P=0.006$), where successful ideas exhibited divergence and unsuccessful ideas exhibited convergence of semantic similarity (Fig. 3A). The information content manifested a trend toward significant crossover interaction ($F_{1,11}=4.0$, $P=0.072$), where successful ideas increased and unsuccessful ideas decreased their information content in time (Fig. 3B). The polysemy exhibited crossover interaction decreasing in time for successful ideas ($F_{1,11}=12.8$, $P=0.004$; Fig. 3C), whereas the average level of abstraction decreased in time but with only a trend toward significance ($F_{1,11}=4.6$, $P=0.055$; Fig. 3D). Because design practice usually generates both successful and unsuccessful ideas, these results support models of concurrent divergent ideation and convergent evaluation in creative problem solving.

### 3.3. *IC-based semantic similarity measures outperform path-based ones*

The majority of 40 different semantic similarity formulas generated highly correlated outputs, which segregated them into clusters of purely IC-based, hybrid path/IC-based, and path-based similarity measures (Fig. 4). Motivated by the significant difference detected in the time dynamics of semantic similarity between successful and unsuccessful ideas, we performed post hoc linear contrasts in individual two-factor rANOVAs and ranked the 40 semantic similarity measures by the observed statistical power (Fig. 5; Table 2). The best performance was achieved by purely IC-based similarity measures using the formulas by Lin ($F_{1,11}>10.6$, $P<0.008$, power>0.84), Jiang–Conrath ($F_{1,11}>10.2$, $P<0.008$, power>0.83), and Resnik ($F_{1,11}>8.3$, $P<0.015$, power>0.75; Fig. 5), all of which rely on the calculation of the lowest common subsumer of pairs of nouns. Hybrid path/IC-based similarity measures had a weaker performance as exemplified by the Meng formula for all IC measures ($F_{1,11}>6.6$, $P<0.026$, power>0.65), and the Zhou formula ($F_{1,11}>6.0$, $P<0.031$, power>0.61) for all IC measures except IC Sanchez for which there was only a trend toward significance ($F_{1,11}=4.3$, $P=0.063$, power=0.47). Path-based similarity measures underperformed ($F_{1,11}>4.4$, $P<0.059$, power=0.48), even though the Wu–Palmer ($F_{1,11}=5.7$, $P=0.035$, power=0.59) and Li ($F_{1,11}=5.1$, $P=0.045$, power=0.54; Fig. 5) measures were statistically significant. Among the IC formulas, the best overall performance was achieved by the cluster of Sanchez–Batet, Blanchard and Seco, which exhibited highly correlated IC values (r>0.93, P<0.001; Fig. 6).

Having ranked the IC formulas (Fig. 5), we also performed individual two-factor rANOVAs for each of the 7 IC measures. The information content of nouns increased/decreased in time for successful/unsuccessful ideas exhibiting a crossover interaction as shown by IC Sanchez–Batet ($F_{1,11}$=6.2, P=0.03), with 4 other IC measures by Blanchard, Meng, Seco and Zhou manifesting a trend toward significance ($F_{1,11}$>3.8, P<0.076). Because the first-ranked IC measure by Sanchez–Batet was significantly changed in the post hoc tests, we interpreted the trend-like significance from the corresponding three-factor rANOVA as a Type II error due to inclusion in the analysis of IC measures that compound the word information content with path-based information (such as the depth of the word in the taxonomy).

### 3.4. *Effect of first feedback on creative problem solving*

Further, to test whether the first feedback from the client influences problem solving, we compared the conversations containing ideas before and after first feedback. Apart from polysemy, which exhibited an interaction between idea type and time ($F_{1,11}$=6.1, P=0.031; Fig. 7C), none of the other 40 semantic similarity measures (two-factor rANOVAs: $F_{1,11}$<2.7, P>0.13; Fig. 7A), 7 information content measures (two-factor rANOVAs: $F_{1,11}$<1.6, P>0.23; Fig. 7B), or the level of abstraction ($F_{1,11}$<0.1, P>0.78; Fig. 7D) differed before and after first feedback when both successful and unsuccessful ideas in the conversations are analyzed together. However, when only the time dynamics of successful ideas is considered, the first feedback led to a divergence of semantic similarity (three-factor rANOVA: $F_{1,6}$=22.8, P=0.003; Fig. 8A) and enhanced the information content (three-factor rANOVA: $F_{1,6}$=6.5, P=0.044; Fig. 8B). Post hoc two-factor rANOVAs found significant differences in 36 of 40 semantic similarity measures ($F_{1,6}$>6.2, P<0.047 for 36 measures; $F_{1,6}$>12.6, P<0.012 for 33 measures excluding Zhou similarity) and in 4 of 7 IC measures by Sanchez–Batet ($F_{1,6}$=25.2, P=0.002), Meng ($F_{1,6}$=11.3, P=0.015), Zhou ($F_{1,6}$=10.4, P=0.018), and Yuan ($F_{1,6}$=6.6, P=0.042), with a trend toward significance for 2 other IC measures by Blanchard ($F_{1,6}$=5.4, P=0.059) and Seco ($F_{1,6}$=5.6, P=0.056). The first feedback also decreased polysemy ($F_{1,6}$=8.2, P=0.029; Fig. 8C) and the average level of abstraction ($F_{1,6}$=16.8, P=0.006; Fig. 8D). These results show that the first feedback from the client has positively altered the process of problem solving and suggest that creativity benefits from external criticism obtained during the time in which the generated ideas are still evolving.

### *3.5. Effect of first evaluation on creative problem solving*

Ideas before first evaluation are subject to change, with new features added and initial features omitted, whereas ideas after first evaluation do not change their main features, only their details. Considering this, we also tested the effects of first evaluation by client or instructor upon problem solving. Conversations containing both successful and unsuccessful ideas before and after first evaluation did not exhibit different time dynamics in any of the 40 semantic similarity measures (two-factor rANOVAs: $F_{1,10}<2.7$, $P>0.14$; Fig. 9A), in any of the 7 information content measures (two-factor rANOVAs: $F_{1,10}<0.9$, $P>0.38$; Fig. 9B), polysemy ($F_{1,10}<3.8$, $P>0.08$; Fig. 9C), or the average level of abstraction ($F_{1,10}<0.1$, $P>0.76$; Fig. 9D). Analyzing the time dynamics of only successful ideas also showed a lack of effect upon 39 of 40 semantic similarity measures (three-factor rANOVA: $F_{1,7}=2.9$, $P=0.131$; two-factor rANOVAs: $F_{1,7}<4.9$, $P>0.063$; Fig. 10A), 7 information content measures (three-factor rANOVA: $F_{1,7}=3.1$, $P=0.124$; two-factor rANOVAs: $F_{1,7}<4.7$, $P>0.067$; Fig. 10B), polysemy ($F_{1,7}=3.8$, $P=0.093$; Fig. 10C), and the average level of abstraction ($F_{1,7}=5.0$, $P=0.06$; Fig. 10D). Only the semantic similarity measure by Rada showed an enhanced divergence after first evaluation ($F_{1,7}=6.0$, $P=0.044$), but we interpreted this as a Type I error since the path-based similarity measures were the weakest in terms of statistical power (Fig. 5). These results suggest that the first evaluation had a minimal effect upon those ideas that were not dropped but developed further.

## 4. Discussion

### *4.1. Implications for cognitive science of creativity*

The presented findings advance cognitive science by showing that convergence and divergence of semantic similarity, as well as time dynamics of information content, polysemy, and level of abstraction, could be evaluated objectively for problem-solving conversations in academic settings and be used to monitor the probability of success of different ideas that are generated and developed in the process of problem solving in view of improving student training, creative thinking and skill acquisition. The observed convergence of semantic similarity for unsuccessful ideas and divergence for successful ideas parallel the psychological definitions of convergent/divergent thinking that associate creativity with divergent thought [3-5]. Thus, the convergence or divergence of semantic similarity in verbalized thoughts could be interpreted as a faithful reflection of the underlying cognitive processes, including convergent (analytical) or divergent (associative) thinking. Given the correspondence between convergence/divergence of semantic similarity and convergent/divergent thinking, our results, with regard to successful/unsuccessful ideas, provide extra support to recent accounts of concurrent occurrence of convergent and divergent thinking in creative problem solving [12, 15, 19].

Psychological accounts of creative thinking and problem solving describe divergent generation of novelty and convergent exploration, evaluation or elimination of the introduced novelty [19]. The opposite trend line slopes for successful and unsuccessful ideas found in the studied design review conversations can be well explained by difference in the rates of divergent production and convergent elimination of novelty. Thus, convergent (analytical) and divergent (associative) cognitive processes, quantified through time dynamics of semantic similarity, appear to be the main factors that shape the evolution and determine the outcome of generated ideas.

Language is a powerful data source for the analysis of mental processes, such as design and creative problem solving. Extracting meaningful results about the cognitive processes underlying human creativity from recorded design conversations, however, is a challenging task because not all aspects of human creative skills are verbalized or represented at a consciously accessible level [25]. Semantic networks address the latter problem by providing a structured representation of not only the explicitly verbalized concepts contained in the conversations [26, 27], but also of the inexplicitly imaged virtual concepts (connecting the verbalized concepts), which are extracted from available lexical databases that are independent of the designer's background [51]. In our methods, we have used WordNet 3.1 as a lexical database and have constructed semantic networks containing only nouns. Working with a single lexical category (nouns) was necessitated by the fact that WordNet consists of four subnets, one each for nouns, verbs, adjectives, and adverbs, with only a few cross-subnet pointers [22]. Besides nouns being the largest and deepest hierarchical taxonomy in WordNet, our choice to construct semantic networks of nouns had been motivated by previous findings that showed how: noun phrases are useful surrogates for measuring early phases of the mechanical design process in educational settings [52], networks of nouns act as stimuli for idea generation in creative problem solving [53], noun-noun combinations and noun-noun relations play essential role in designing [54], and similarity/dissimilarity of noun-noun combinations is related to creativity through yielding emergent properties of generated ideas [55]. Noteworthy, disambiguation of noun senses is not done for the construction of semantic networks because nouns used to describe creative design ideas may acquire new senses different from dictionary-defined ones and polysemy may be responsible for the association of ideas previously thought to be unrelated [56, 57]. The effectiveness of semantic networks of nouns for constructive simulation of difficult-to-observe design-thinking processes and investigation of creativity in conceptual design was validated in previous studies using different sets of experimental data [26, 27, 51, 58].

The temporal factor is not a prerequisite for applying semantic network analysis to text data, however, determining the slope of convergence/divergence is essential if the objective is to understand dynamic processes or to achieve dynamic control of artificial intelligence applications. The temporal resolution of the method for studying cognitive processes in humans is limited by the speed of verbalization and the sparsity of nouns in the sentences. A possible inclusion of more lexical categories in the semantic analysis would increase the temporal resolution by allowing verbal reports to be divided into smaller pieces of text, but for practical realization this will require further extensive information theoretic research on how semantic similarity could be meaningfully defined for combinations of lexical categories, such as verbs and nouns, which form separate hierarchical taxonomies in WordNet.

### *4.2.      Implications for artificial intelligence research*

Implementing creativity in machines endowed with artificial intelligence requires mechanisms for generation of conceptual space within which creative activity occurs and algorithms for exploration or transformation of the conceptual space [59]. The most serious challenge, however, is considered not the production of novel ideas, but their automated evaluation [50]. For example, machines could explore structured conceptual spaces and combine or transform ideas in new ways, but then arrive at solutions that are of no interest or value to humans. Since creativity requires both novelty and a positive evaluation of the product, the engineering of creative machines is conditional on the availability of algorithms that could compute the poor quality of newly generated ideas, thereby allowing ideas to be dropped or amended accordingly [50].

Linkography is a method for analyzing decisions and activities that occur during a design work session by parsing the design conversations into a large number of small steps called design moves, some of which are then interrelated through backlinks to previous moves or forelinks to future moves. The most significant elements in a linkograph are critical moves, which are particularly rich in links. The percentage of critical moves and the link index (the ratio between the number of links and the number of moves) are positively correlated with creativity. The ideas considered most meaningful (successful ideas) have a significantly higher number of links than other ideas [60]. Information theoretic approach to measuring creativity in linkography has further shown that the Shannon entropy $H$ of the linkograph is not directly correlated to the design outcome, however, the slope of the rate of change in entropy (second derivative in time of the entropy curve, $\frac{d^2H}{dt^2}$) for high-scoring design sessions (successful ideas) is positive, whereas for low-scoring design sessions (unsuccessful ideas) is negative [61].

Here, we have analyzed design review conversations at the level of individual words and extracted nouns from the corresponding text transcripts through computer automated natural language processing. With the use of semantic networks of nouns constructed at different times, we studied the time dynamics of 49 semantic measures that quantitatively evaluated the content of generated ideas in creative problem solving. We found that the creative ideas, which are judged as successful by the client, exhibit distinct dynamics including divergence of semantic similarity, increased information content and decreased polysemy in time. These findings are susceptible to reverse-engineering and could be useful for the development of machines endowed with general artificial intelligence that are capable of using language (words) and abstract concepts (meanings) to assist in solving problems that are currently reserved only for humans [62]. A foreseeable application would be to use divergence of Lin/Sanchez–Batet semantic similarity in computer-assisted enhancement of human creativity wherein a software proposes a set of possible solutions or transformations of generated ideas and the human designer chooses which of the proposed ideas to drop and which to transform further. As an example, consider a design task described by the set of nouns { bird, crayon, desk, hand, paper } whose average semantic similarity is 0.39. The software computes four possible solutions that change the average similarity of the set when added to it, namely, drawing (0.40), sketch (0.39), greeting_card (0.35), origami (0.29), and proposes origami as the most creative solution as it is the most divergent. If the designer rejects the idea, the software proposes greeting_card as the second best choice, and so on. Divergence of semantic similarity could be monitored and used to supplement existing systems for support of user creativity [63-65]. Accumulated experience with software that enhances human creativity could help optimize the evaluation function for dynamic transformation of semantic similarity and information content of generated ideas up to the point wherein the computer-assisted design products are invariably more successful than products designed without computer aid. If such an optimized evaluation function is arrived at, creative machines could be able to evaluate their generated solutions at different stages without human help, and steer a selected design solution toward success through consecutive transformations; human designers would then act as clients who run design tasks with slightly different initial constraints on the design problem and at the end choose the computer product that best satisfies their personal preference.

### *4.3.     Future work*

Having established a method for the quantitative evaluation of convergence/divergence in creative problem solving and design, we are planning to utilize it for the development of artificial intelligence applications, the most promising of which are software for the computer-assisted enhancement of human creativity and bot-automated design education in massive open online courses (MOOCs), wherein a few instructors are assisted by artificial agents that provide feedback on the design work for thousands of students. We are also interested in cross-validating our results with the use of conversation transcripts from the design process of professional design teams in which the instructor-student paradigm is not applicable, and testing whether semantic measure analysis of online texts in social media or social networks could predict future human behavior.

### *Ethics statement*

The authors have signed Data-Use Agreements to Dr. Robin Adams (Purdue University) for accessing the Purdue DTRS Design Review Conversations Database, thereby agreeing not to reveal personal identifiers in published results and not to create any commercial products.

**Acknowledgement**

G.V.G. acknowledges partial financial support for this study from Grant-in-Aid 25750001 by the Japan Society for the Promotion of Science (JSPS).

**Figures**

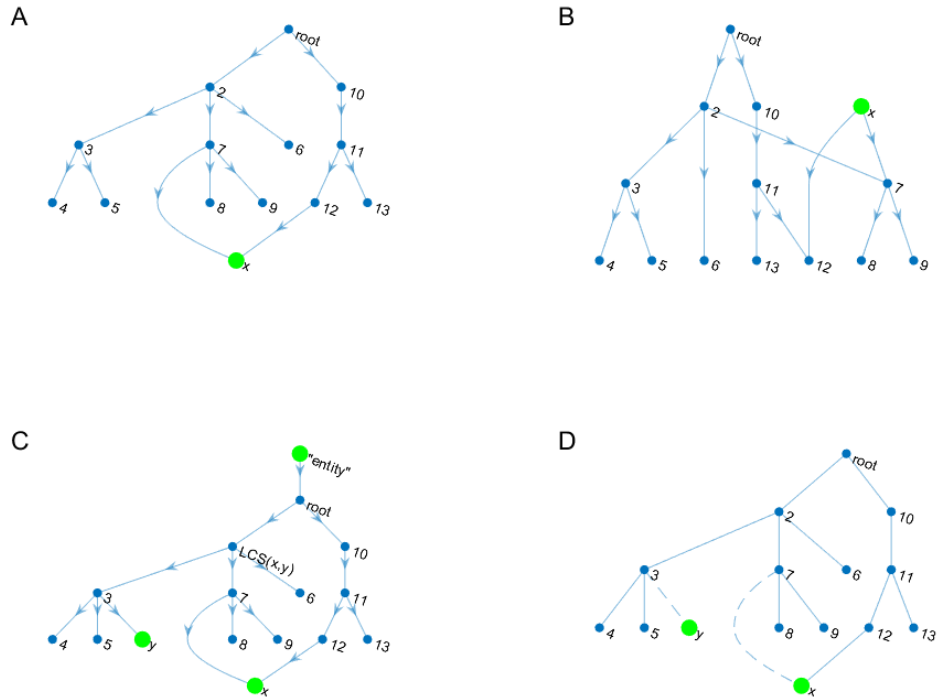

*Fig. 1. Calculation of semantic functions that take word arguments in WordNet 3.1 fragment composed of meaning vertices (blue) and word vertices (green). (A) $\text{Polysemy}(x) = 2$ ; $\text{Depth}(x) = 3$ ; $|\text{Subsumers}(x)| = 6$ . (B) $|\text{Subvertices}(x)| = 4$ ; $\text{Leaves}(x) = \{8, 9, 12\}$ ; $|\text{Leaves}(x)| = 3$ ; $\text{Commonness}(x) = 3/4$ . (C) $\text{LCS}(x, y)$ ; $\text{Depth}[\text{LCS}(x, y)] = 2$ . (D) $\text{Distance}(x, y) = 2$ excludes the dashed edges that connect the words $x$ and $y$ to their meaning vertices.*

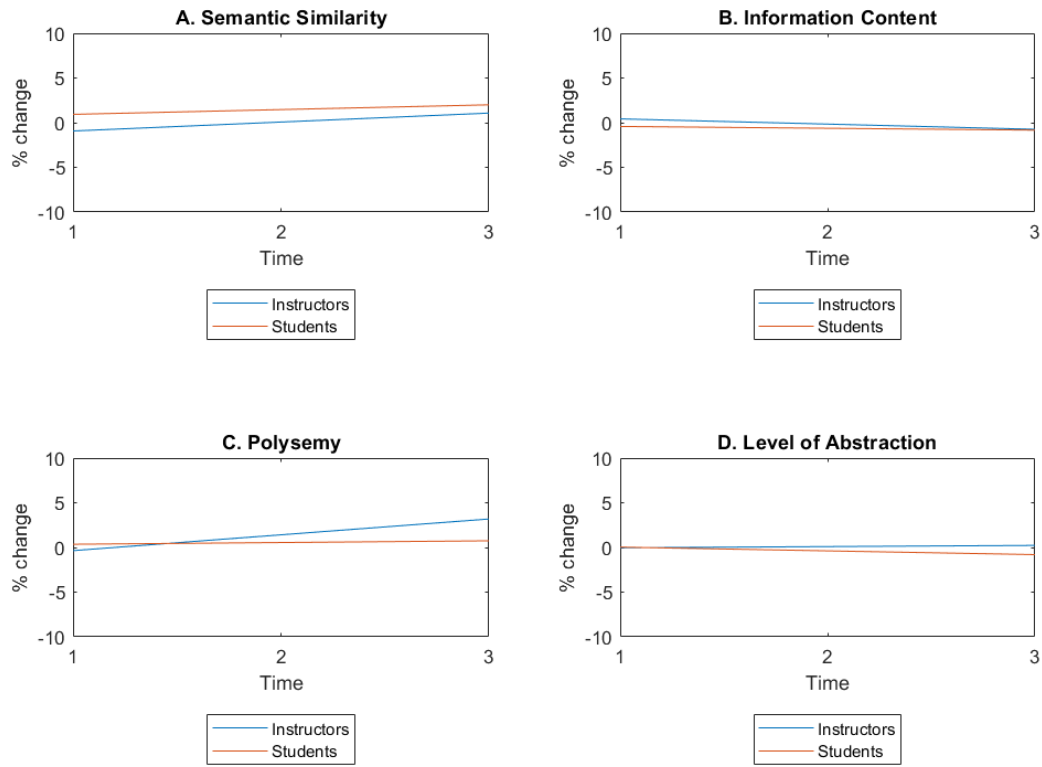

*Fig. 2. Comparison between student thinking and instructor thinking. Linear trend lines show the time dynamics of semantic similarity (average of 40 measures) (A), information content (average of 7 measures) (B), polysemy (C), and the level of abstraction (D) of nouns in semantic networks constructed from design problem-solving conversations.*

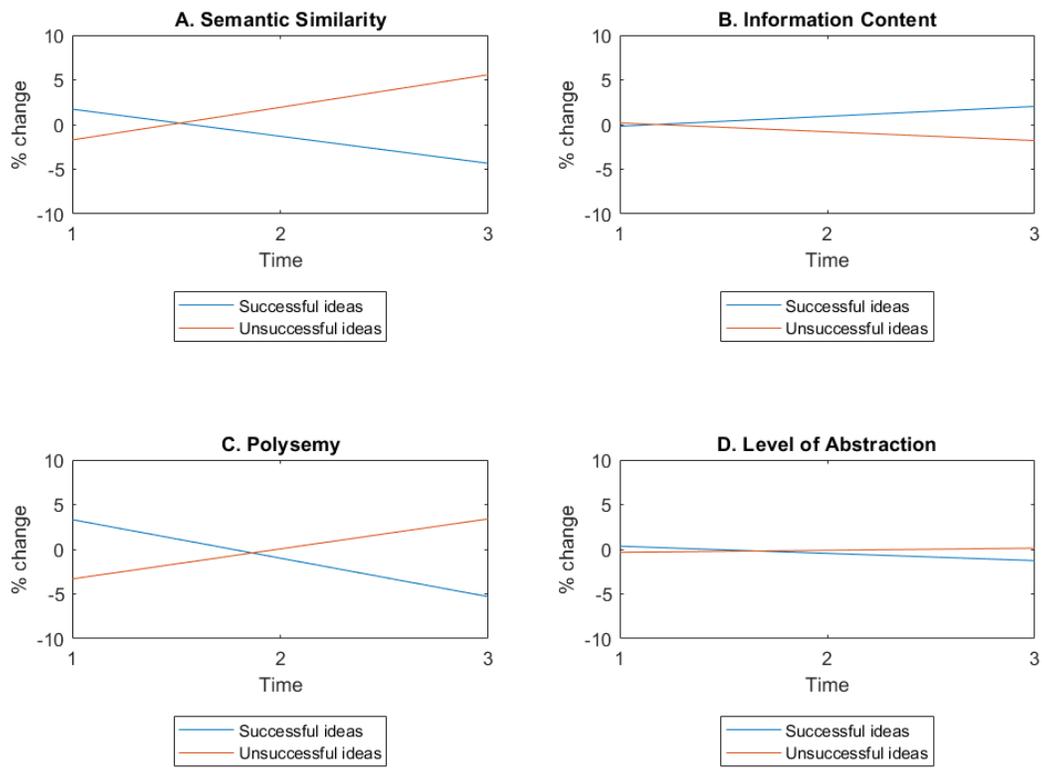

*Fig. 3. Comparison of conversations pertaining to successful ideas or unsuccessful ideas.*

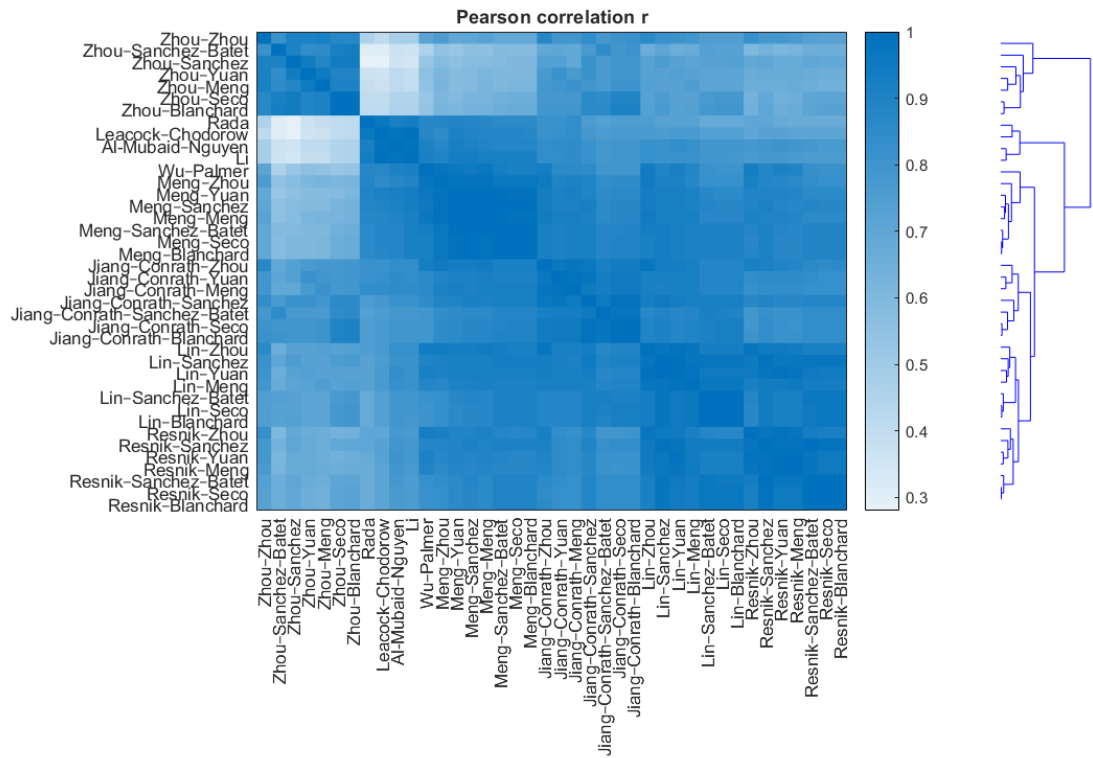

*Fig. 4. Hierarchical clustering of 40 semantic similarity measures based on WordNet 3.1.*

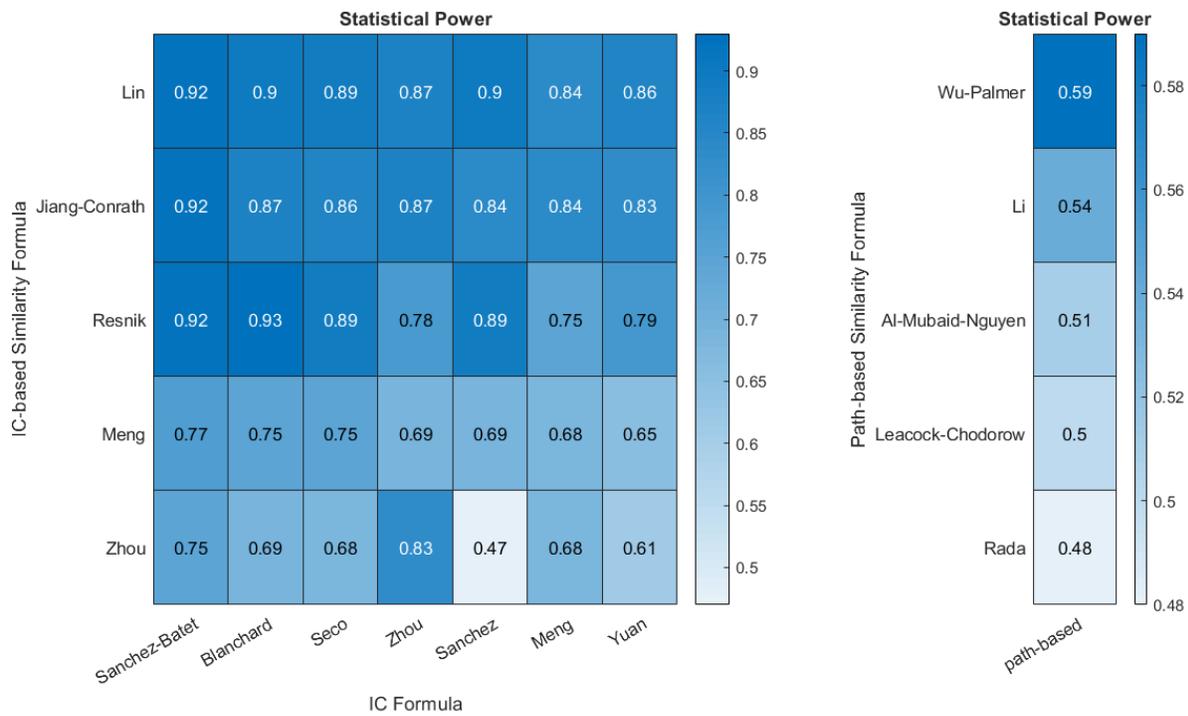

*Fig. 5. Observed statistical power in detecting difference of time dynamics between successful and unsuccessful ideas. IC-based similarity measures outperform path-based similarity measures.*

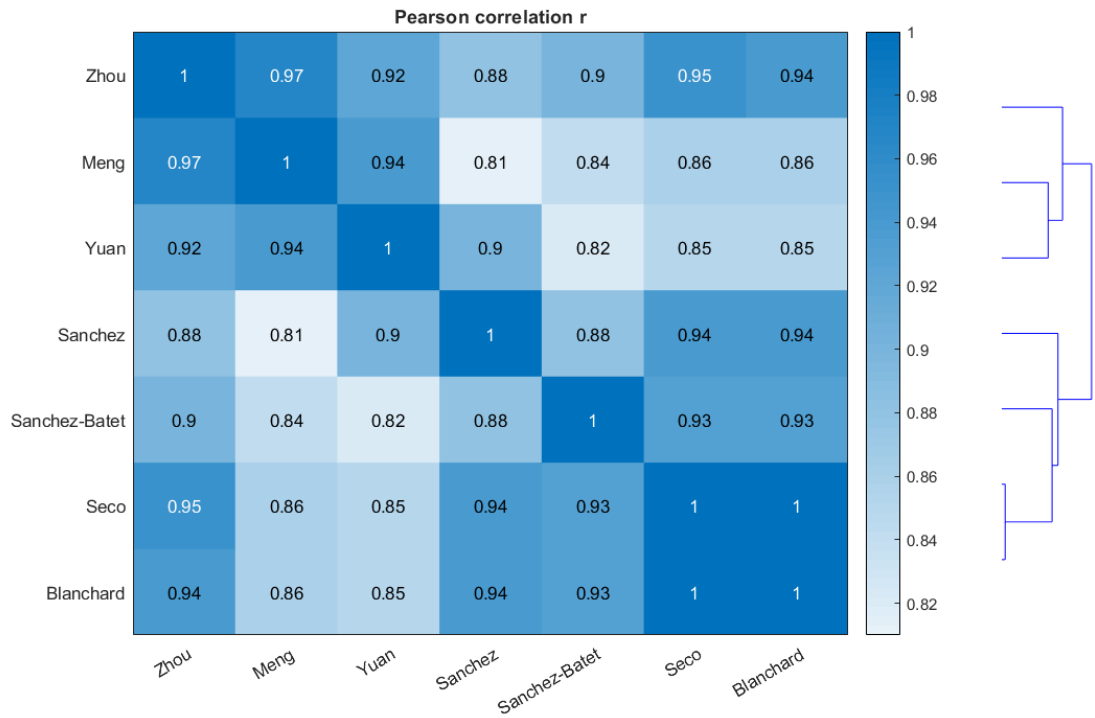

*Fig. 6. Hierarchical clustering of 7 information content (IC) measures based on WordNet 3.1.*

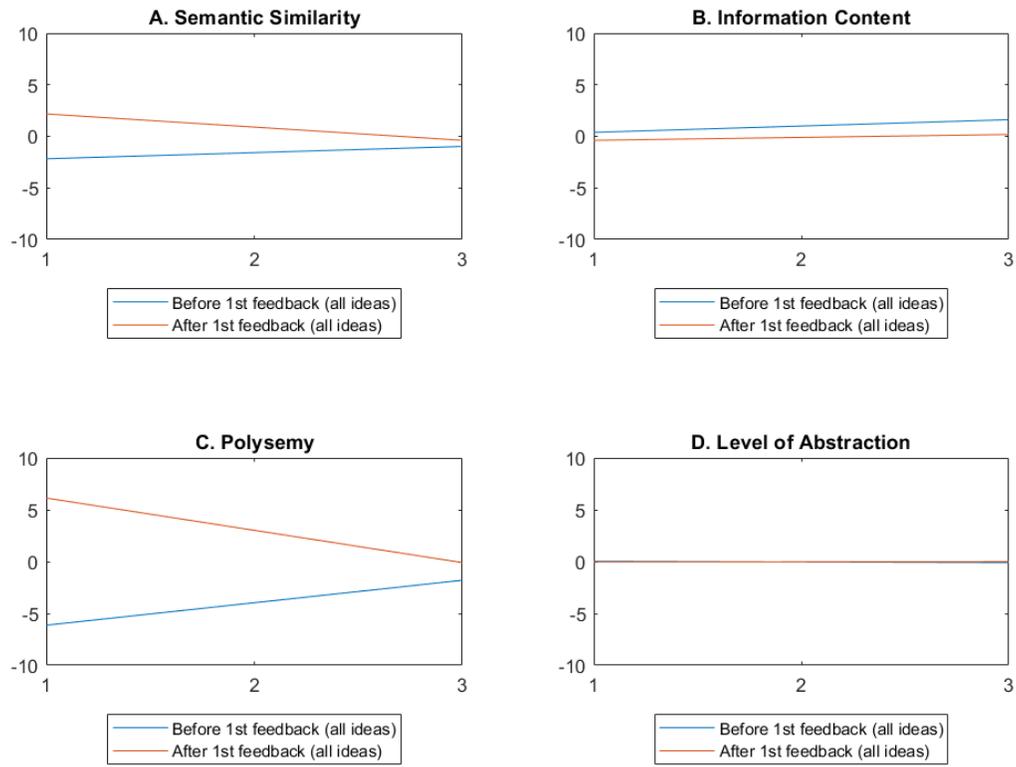

*Fig. 7. Effect of the first feedback by the client on the time dynamics of semantic measures in the conversations containing both successful and unsuccessful ideas.*

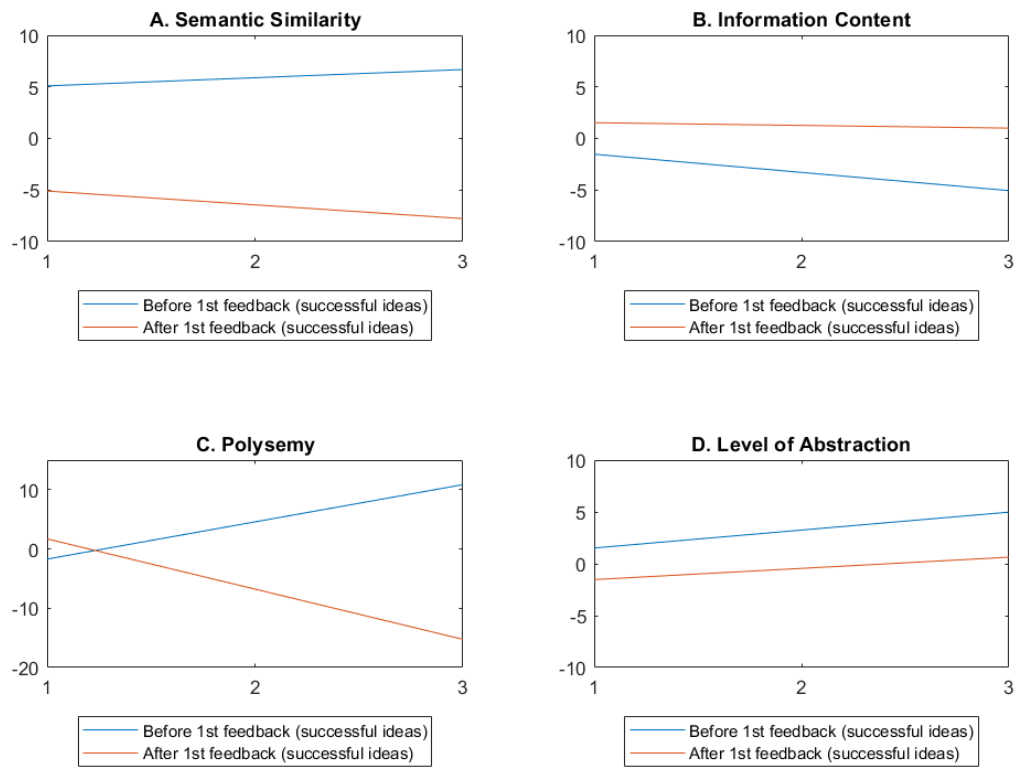

*Fig. 8. Effect of the first feedback by the client on the time dynamics of semantic measures in the conversations containing only successful ideas.*

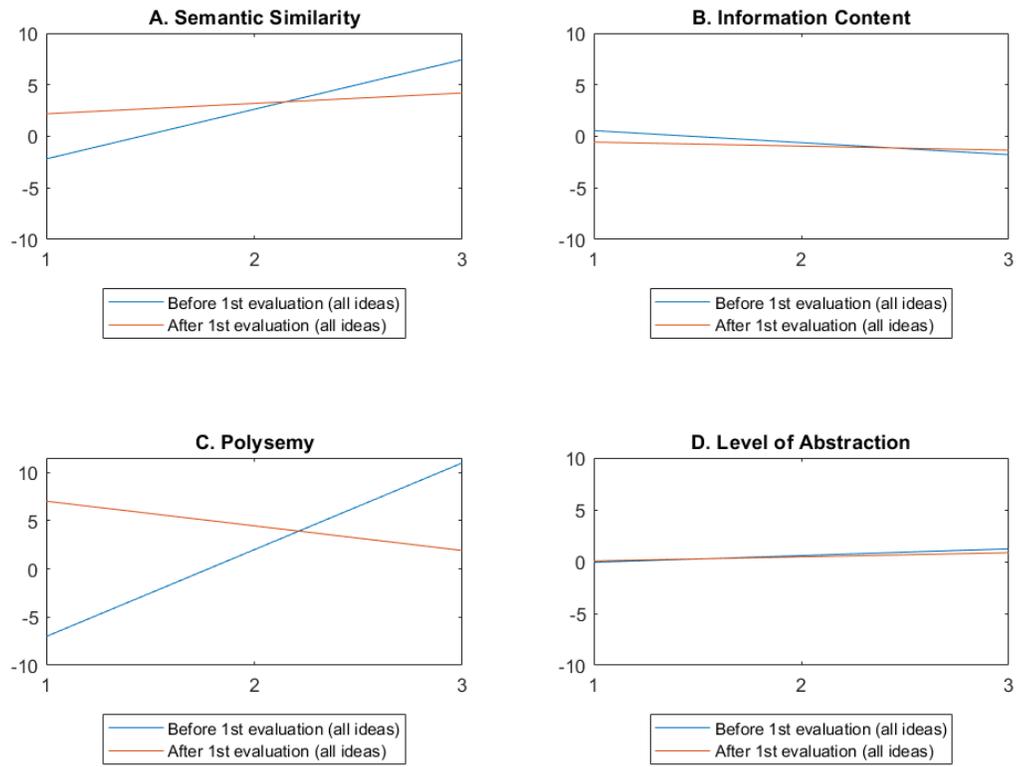

*Fig. 9. Effect of the first evaluation by the instructor or the client on the time dynamics of semantic measures in the conversations containing both successful and unsuccessful ideas.*

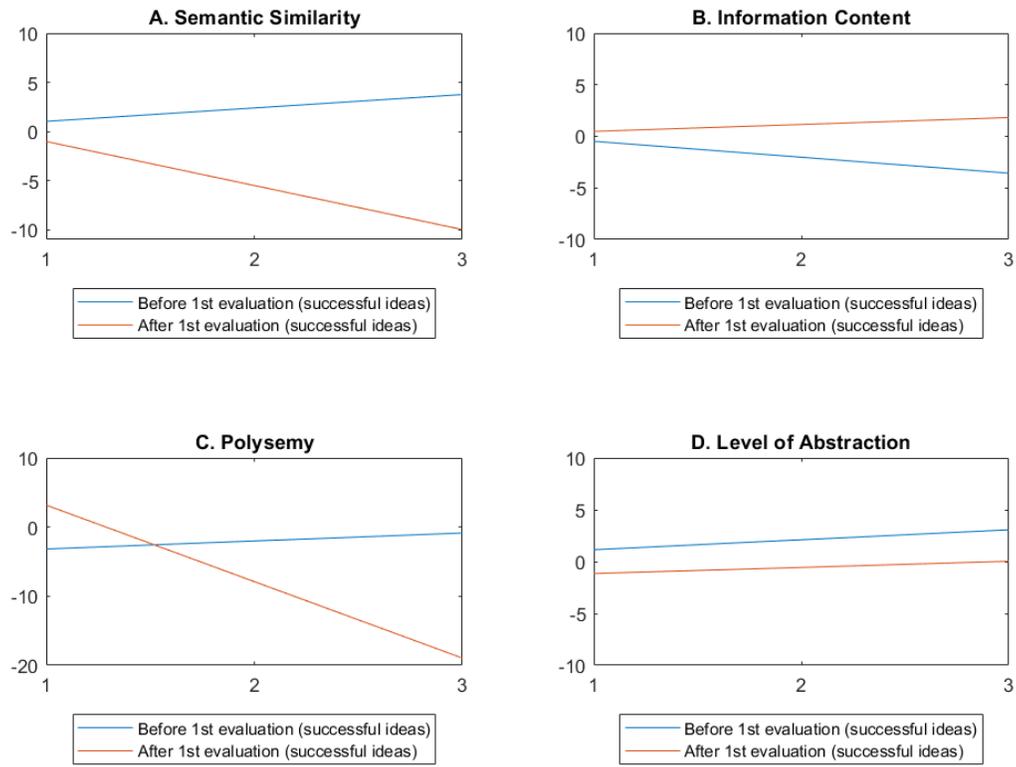

*Fig. 10. Effect of the first evaluation by the instructor or the client on the time dynamics of semantic measures in the conversations containing only successful ideas.*

# References

[1] Y. Wang, In search of cognitive foundations of creativity, in: E.G. Carayannis (Ed.), Encyclopedia of Creativity, Invention, Innovation and Entrepreneurship, Springer, New York, 2013, pp. 902-913.
[2] J.P. Guilford, Creative abilities in the arts, Psychological Review 64(2) (1957) 110-118.
[3] M.A. Runco, Creativity: Theories and Themes: Research, Development, and Practice, Elsevier Academic Press, Burlington, Massachusetts, 2007.
[4] L. Gabora, Revenge of the "neurds": Characterizing creative thought in terms of the structure and dynamics of memory, Creativity Research Journal 22(1) (2010) 1-13.
[5] P.T. Sowden, A. Pringle, L. Gabora, The shifting sands of creative thinking: Connections to dual-process theory, Thinking & Reasoning 21(1) (2015) 40-60.
[6] J.P. Guilford, Three faces of intellect, American Psychologist 14(8) (1959) 469-479.
[7] L. Hudson, Contrary Imaginations: A Psychological Study of the English Schoolboy, Penguin Books, Harmondsworth, 1974.
[8] R. Riding, I. Cheema, Cognitive styles—an overview and integration, Educational Psychology 11(3-4) (1991) 193-215.
[9] J.S.B.T. Evans, Dual-processing accounts of reasoning, judgment, and social cognition, Annual Review of Psychology 59(1) (2008) 255-278.
[10] S.M. Smith, T.B. Ward, R.A. Finke, The creative cognition approach, MIT Press, Cambridge, Massachusetts, 1995.
[11] R. Fischer, B. Hommel, Deep thinking increases task-set shielding and reduces shifting flexibility in dual-task performance, Cognition 123(2) (2012) 303-307.
[12] B.R. Lawson, Parallel lines of thought, Languages of Design 1(4) (1993) 357-366.
[13] J. Chan, C.D. Schunn, The importance of iteration in creative conceptual combination, Cognition 145 (2015) 104-115.
[14] B.R. Lawson, How Designers Think: The Design Process Demystified, Architectural Press, Oxford, 2005.
[15] G. Goldschmidt, Linkographic evidence for concurrent divergent and convergent thinking in creative design, Creativity Research Journal 28(2) (2016) 115-122.
[16] D.R. Brophy, Comparing the attributes, activities, and performance of divergent, convergent, and combination thinkers, Creativity Research Journal 13(3-4) (2001) 439-455.
[17] M. Basadur, Optimal ideation-evaluation ratios, Creativity Research Journal 8(1) (1995) 63-75.
[18] N. Cross, Engineering Design Methods: Strategies for Product Design, Third Edition ed., John Wiley & Sons, Chichester, UK, 2000.
[19] A. Cropley, In praise of convergent thinking, Creativity Research Journal 18(3) (2006) 391-404.
[20] A.-M. Olteţeanu, Z. Falomir, comRAT-C: A computational compound Remote Associates Test solver based on language data and its comparison to human performance, Pattern Recognition Letters 67(1) (2015) 81-90.
[21] R.S. Adams, J.A. Siddiqui, Purdue DTRS – Design Review Conversations Database, XRoads Technical Report, TR-01-13, Purdue University, West Lafayette, Indiana, 2013.
[22] C. Fellbaum, WordNet: An Electronic Lexical Database, The MIT Press, Cambridge, Massachusetts, 1998.

[23] K.B. Clark, T. Fujimoto, The power of product integrity, Harvard Business Review 68(6) (1990) 107-118.
[24] K.T. Ulrich, S.D. Eppinger, Product Design and Development, 6th ed., McGraw-Hill, New York, 2016.
[25] M.A. Boden, The Creative Mind: Myths and Mechanisms, Second ed., Routledge, London, 2004.
[26] G.V. Georgiev, Y. Nagai, T. Taura, Method of design evaluation focused on relations of meanings for a successful design, in: D. Marjanovic, M. Storga, N. Pavkovic, N. Bojcetic (Eds.) 10th International Design Conference, DESIGN 2008, The Design Society, Dubrovnik, Croatia, 2008, pp. 1149-1158.
[27] G.V. Georgiev, Y. Nagai, T. Taura, A method for the evaluation of meaning structures and its application in conceptual design, Journal of Design Research 8(3) (2010) 214-234.
[28] S. Bird, E. Klein, E. Loper, Natural Language Processing with Python, O'Reilly Media, Sebastopol, California, 2009.
[29] S. Loria, TextBlob: Simplified Text Processing, Center for Open Science, Charlottesville, Virginia, 2016.
[30] L. Meng, R. Huang, J. Gu, A review of semantic similarity measures in WordNet, International Journal of Hybrid Information Technology 6(1) (2013) 1-12.
[31] E. Blanchard, M. Harzallah, P. Kuntz, A generic framework for comparing semantic similarities on a subsumption hierarchy, in: M. Ghallab, C.D. Spyropoulos, N. Fakotakis, N. Avouris (Eds.), ECAI 2008: 18th European Conference on Artificial Intelligence including Prestigious Applications of Intelligent Systems (PAIS 2008), IOS Press, Patras, Greece, 2008, pp. 20-24.
[32] L. Meng, J. Gu, Z. Zhou, A new model of information content based on concept's topology for measuring semantic similarity in WordNet, International Journal of Grid and Distributed Computing 5(3) (2012) 81-94.
[33] D. Sánchez, M. Batet, D. Isern, Ontology-based information content computation, Knowledge-Based Systems 24(2) (2011) 297-303.
[34] D. Sánchez, M. Batet, A new model to compute the information content of concepts from taxonomic knowledge, International Journal on Semantic Web & Information Systems 8(2) (2012) 34-50.
[35] N. Seco, T. Veale, J. Hayes, An intrinsic information content metric for semantic similarity in WordNet, in: R.L. de Mántaras, L. Saitta (Eds.), ECAI 2004: 16th European Conference on Artificial Intelligence including Prestigious Applicants of Intelligent Systems (PAIS 2004), IOS Press, Valencia, Spain, 2004, pp. 1089-1090.
[36] Q. Yuan, Z. Yu, K. Wang, A new model of information content for measuring the semantic similarity between concepts, 2013 International Conference on Cloud Computing and Big Data, Institute of Electrical and Electronics Engineers2013, pp. 141-146.
[37] Z. Zhou, Y. Wang, J. Gu, A new model of information content for semantic similarity in WordNet, FGCNS '08 Second International Conference on Future Generation Communication and Networking Symposia, Institute of Electrical and Electronics Engineers2008, pp. 85-89.
[38] H. Al-Mubaid, H.A. Nguyen, A cluster-based approach for semantic similarity in the biomedical domain, 2006 International Conference of the IEEE Engineering in Medicine and Biology Society, Institute of Electrical and Electronics Engineers2006, pp. 2713-2717.

[39] C. Leacock, M. Chodorow, Combining local context and WordNet similarity for word sense identification, in: C. Fellbaum (Ed.), WordNet: An Electronic Lexical Database, MIT Press, Cambridge, Massachusetts, 1998, pp. 265-283.
[40] Y. Li, Z.A. Bandar, D. McLean, An approach for measuring semantic similarity between words using multiple information sources, IEEE Transactions on Knowledge and Data Engineering 15(4) (2003) 871-882.
[41] R. Rada, H. Mili, E. Bicknell, M. Blettner, Development and application of a metric on semantic nets, IEEE Transactions on Systems, Man, and Cybernetics 19(1) (1989) 17-30.
[42] Z. Wu, M. Palmer, Verbs semantics and lexical selection, 32nd Annual Meeting of the Association for Computational Linguistics, Association for Computational Linguistics, Las Cruces, New Mexico, 1994, pp. 133-138.
[43] J.J. Jiang, D.W. Conrath, Semantic similarity based on corpus statistics and lexical taxonomy, International Conference on Research on Computational Linguistics (ROCLING X), Association for Computational Linguistics and Chinese Language Processing, Taipei, Taiwan, 1997, pp. 19-33.
[44] D. Lin, An information-theoretic definition of similarity, ICML '98 Proceedings of the 15th International Conference on Machine Learning, Morgan Kaufmann Publishers1998, pp. 296-304.
[45] L. Meng, R. Huang, J. Gu, Measuring semantic similarity of word pairs using path and information content, International Journal of Future Generation Communication and Networking 7(3) (2014) 183-194.
[46] P. Resnik, Using information content to evaluate semantic similarity in a taxonomy, IJCAI'95 Proceedings of the 14th international joint conference on Artificial intelligence - Volume 1, Morgan Kaufmann Publishers1995, pp. 448-453.
[47] Z. Zhou, Y. Wang, J. Gu, New model of semantic similarity measuring in Wordnet, ISKE 2008, 3rd International Conference on Intelligent System and Knowledge Engineering, Institute of Electrical and Electronics Engineers2008, pp. 256-261.
[48] C.P. Doncaster, A.J.H. Davey, Analysis of Variance and Covariance: How to Choose and Construct Models for the Life Sciences, Cambridge University Press, Cambridge, 2007.
[49] T. Dartnall, Artificial Intelligence and Creativity: An Interdisciplinary Approach, Springer, Dordrecht, 1994.
[50] M.A. Boden, Creativity and artificial intelligence, Artificial Intelligence 103(1) (1998) 347-356.
[51] E. Yamamoto, M. Goka, N.F.M. Yusof, T. Taura, Y. Nagai, Virtual modeling of concept generation process for understanding and enhancing the nature of design creativity, in: M. Norell Bergendahl, M. Grimheden, L. Leifer, P. Skogstad, U. Lindemann (Eds.) 17th International Conference on Engineering Design, Vol. 2, Design Theory and Research Methodology, The Design Society, Palo Alto, CA, 2009, pp. 101-112.
[52] A. Mabogunje, L.J. Leifer, Noun phrases as surrogates for measuring early phases of the mechanical design process, 1997 ASME Design Engineering Technical Conferences: DETC '97, American Society of Mechanical Engineers, Sacramento, California, 1997.
[53] N.M. Segers, B. de Vries, H.H. Achten, Do word graphs stimulate design?, Design Studies 26(6) (2005) 625-647.
[54] A. Dong, The Language of Design: Theory and Computation, Springer, London, 2009.
[55] M.J. Wilkenfeld, T.B. Ward, Similarity and emergence in conceptual combination, Journal of Memory and Language 45(1) (2001) 21-38.

[56] G.V. Georgiev, T. Taura, Polysemy in design review conversations, 10th Design Thinking Research Symposium, Purdue University, Purdue University, West Lafayette, Indiana, 2014.
[57] T. Taura, Y. Nagai, Concept Generation for Design Creativity: A Systematized Theory and Methodology, Springer, London, 2013.
[58] T. Taura, E. Yamamoto, M.Y.N. Fasiha, M. Goka, F. Mukai, Y. Nagai, H. Nakashima, Constructive simulation of creative concept generation process in design: a research method for difficult-to-observe design-thinking processes, Journal of Engineering Design 23(4) (2012) 297-321.
[59] N. Kelly, J.S. Gero, Situated interpretation in computational creativity, Knowledge-Based Systems 80 (2015) 48-57.
[60] G. Goldschmidt, Linkography: Unfolding the Design Process, MIT Press, Cambridge, Massachusetts, 2014.
[61] J.W.T. Kan, J.S. Gero, Quantitative Methods for Studying Design Protocols, Springer, Dordrecht, 2017.
[62] J. McCarthy, M.L. Minsky, N. Rochester, C.E. Shannon, A Proposal for the Dartmouth Summer Research Project on Artificial Intelligence, August 31, 1955, AI Magazine 27(4) (2006) 12-14.
[63] M. Sugimoto, K. Hori, S. Ohsuga, A system to visualize different viewpoints for supporting researchers' creativity, Knowledge-Based Systems 9(6) (1996) 369-376.
[64] Y. Sumi, K. Hori, S. Ohsuga, Supporting the acquisition and modeling of requirements in software design, Knowledge-Based Systems 11(7) (1998) 449-456.
[65] D. Cavaliere, S. Senatore, V. Loia, Context-aware profiling of concepts from a semantic topological space, Knowledge-Based Systems 130 (2017) 102-115.

*Table 1.* Students and design review conversations in the Industrial Design Junior (J) and Graduate (G) subsets. Division of conversations (C1-C5) for comparative analyses (1-4) into groups is indicated as follows: 1a, student; 1b, instructor; 2a, successful; 2b, unsuccessful; 3a, before first feedback; 3b, after first feedback; 4a, before first evaluation; 4b, after first evaluation. For empty cells no video data or transcripts were provided in the dataset.

| Students | C1 | C2 | C3 | C4 | C5 |
|---|---|---|---|---|---|
| J1 |  |  | 1a,b, 2a,b, 3a,b, 4a | 1a,b, 2a,b, 3b, 4a | 1a,b, 2a,b, 3b, 4b |
| J2 |  | 1a,b, 2a,b, 3a, 4a | 1a,b, 2a,b, 3a,b, 4a |  | 1a,b, 2a,b, 3b, 4b |
| J3 |  | 1a,b, 2a,b, 3a, 4a | 1a,b, 2a,b, 3a,b, 4a |  | 1a,b, 2a,b, 3b, 4b |
| J4 |  |  | 1a,b, 2a,b, 3a,b, 4a | 1a,b, 2a,b, 3b, 4a | 1a,b, 2a,b, 3b, 4b |
| J5 | 1a,b, 2a,b, 3a |  | 1a,b, 2a,b, 3a,b |  |  |
| J6 |  | 1a,b, 2a,b, 3a, 4a | 1a,b, 2a,b, 3a,b, 4a | 1a,b, 2a,b, 3b, 4a | 1a,b, 2a,b, 3b, 4b |
| J7 | 1a,b, 2a,b, 3a, 4a |  | 1a,b, 2a,b, 3a,b, 4a | 1a,b, 2a,b, 3b, 4a | 1a,b, 2a,b, 3b, 4b |
| G1 | 1a,b, 3a | 1a,b, 3a |  |  |  |
| G2 | 1a,b, 3a, 4a | 1a,b, 2a,b, 3a, 4a | 1a,b, 2a,b, 3a,b, 4a,b |  |  |
| G3 | 1a,b, 3a, 4a |  | 1a,b, 2a,b, 3a,b, 4a,b | 1a,b, 2a,b, 3b, 4b |  |
| G4 | 1a,b, 3a, 4a | 1a,b, 2a,b, 3a, 4a | 1a,b, 2a,b, 3a,b, 4a,b |  |  |
| G5 | 1a,b, 3a, 4a | 1a,b, 2a,b, 3a, 4a | 1a,b, 2a,b, 3a,b, 4a,b |  |  |
| G6 |  | 1a,b, 2a,b, 3a, 4a | 1a,b, 2a,b, 3a,b, 4a,b | 1a,b, 2a,b, 3b, 4b |  |

*Table 2. Statistics from the post hoc two-factor rANOVAs (linear contrasts of idea*time interaction) used to rank the 40 semantic similarity measures and trend line parameters (y=kt+b) for successful ideas ($k_1$, $b_1$) and unsuccessful ideas ($k_2$, $b_2$) at 3 time points t={1,2,3} in the conversations.*

|  |  | Information Content (IC) Formula | | | | | | |
|---|---|---|---|---|---|---|---|---|
|  |  | Sanchez–Batet | Blanchard | Seco | Zhou | Sanchez | Meng | Yuan |
| IC-based similarity Formula | Lin | $k_1$= -5.250<br>$b_1$= 8.833<br>$k_2$= 6.841<br>$b_2$= -10.424<br>$F_{1,11}$= 13.539<br>P= 0.004 | $k_1$= -4.851<br>$b_1$= 7.604<br>$k_2$= 5.954<br>$b_2$= -8.707<br>$F_{1,11}$= 12.682<br>P= 0.004 | $k_1$= -4.841<br>$b_1$= 7.510<br>$k_2$= 5.960<br>$b_2$= -8.629<br>$F_{1,11}$= 12.360<br>P= 0.005 | $k_1$= -3.495<br>$b_1$= 5.340<br>$k_2$= 4.398<br>$b_2$= -6.242<br>$F_{1,11}$= 11.514<br>P= 0.006 | $k_1$= -3.635<br>$b_1$= 5.771<br>$k_2$= 4.934<br>$b_2$= -7.070<br>$F_{1,11}$= 12.720<br>P= 0.004 | $k_1$= -4.753<br>$b_1$= 6.980<br>$k_2$= 5.334<br>$b_2$= -7.561<br>$F_{1,11}$= 10.615<br>P= 0.008 | $k_1$= -4.059<br>$b_1$= 6.273<br>$k_2$= 5.041<br>$b_2$= -7.255<br>$F_{1,11}$= 11.124<br>P= 0.007 |
| IC-based similarity Formula | Jiang–Conrath | $k_1$= -2.927<br>$b_1$= 4.687<br>$k_2$= 3.301<br>$b_2$= -5.062<br>$F_{1,11}$= 13.428<br>P= 0.004 | $k_1$= -3.139<br>$b_1$= 4.635<br>$k_2$= 3.476<br>$b_2$= -4.973<br>$F_{1,11}$= 11.423<br>P= 0.006 | $k_1$= -3.060<br>$b_1$= 4.424<br>$k_2$= 3.337<br>$b_2$= -4.701<br>$F_{1,11}$= 11.099<br>P= 0.007 | $k_1$= -2.223<br>$b_1$= 3.295<br>$k_2$= 2.445<br>$b_2$= -3.516<br>$F_{1,11}$= 11.459<br>P= 0.006 | $k_1$= -2.292<br>$b_1$= 3.313<br>$k_2$= 2.973<br>$b_2$= -3.994<br>$F_{1,11}$= 10.643<br>P= 0.008 | $k_1$= -2.096<br>$b_1$= 3.060<br>$k_2$= 2.015<br>$b_2$= -2.979<br>$F_{1,11}$= 10.652<br>P= 0.008 | $k_1$= -2.563<br>$b_1$= 3.645<br>$k_2$= 2.728<br>$b_2$= -3.810<br>$F_{1,11}$= 10.241<br>P= 0.008 |
| IC-based similarity Formula | Resnik | $k_1$= -4.309<br>$b_1$= 7.610<br>$k_2$= 5.250<br>$b_2$= -8.550<br>$F_{1,11}$= 13.728<br>P= 0.003 | $k_1$= -4.309<br>$b_1$= 7.591<br>$k_2$= 5.165<br>$b_2$= -8.446<br>$F_{1,11}$= 14.027<br>P= 0.003 | $k_1$= -4.276<br>$b_1$= 7.408<br>$k_2$= 5.213<br>$b_2$= -8.346<br>$F_{1,11}$= 12.299<br>P= 0.005 | $k_1$= -2.902<br>$b_1$= 5.026<br>$k_2$= 4.119<br>$b_2$= -6.242<br>$F_{1,11}$= 9.060<br>P= 0.012 | $k_1$= -3.537<br>$b_1$= 6.212<br>$k_2$= 4.559<br>$b_2$= -7.234<br>$F_{1,11}$= 12.105<br>P= 0.005 | $k_1$= -4.042<br>$b_1$= 6.730<br>$k_2$= 5.094<br>$b_2$= -7.783<br>$F_{1,11}$= 8.357<br>P= 0.015 | $k_1$= -3.778<br>$b_1$= 6.410<br>$k_2$= 4.775<br>$b_2$= -7.406<br>$F_{1,11}$= 9.305<br>P= 0.011 |
| Hybrid path/IC-based similarity Formula | Meng | $k_1$= -4.631<br>$b_1$= 7.362<br>$k_2$= 5.744<br>$b_2$= -8.475<br>$F_{1,11}$= 8.683<br>P= 0.013 | $k_1$= -4.432<br>$b_1$= 6.872<br>$k_2$= 5.486<br>$b_2$= -7.925<br>$F_{1,11}$= 8.461<br>P= 0.014 | $k_1$= -4.448<br>$b_1$= 6.855<br>$k_2$= 5.503<br>$b_2$= -7.910<br>$F_{1,11}$= 8.409<br>P= 0.014 | $k_1$= -3.558<br>$b_1$= 5.590<br>$k_2$= 4.467<br>$b_2$= -6.499<br>$F_{1,11}$= 7.254<br>P= 0.021 | $k_1$= -4.017<br>$b_1$= 6.179<br>$k_2$= 5.008<br>$b_2$= -7.169<br>$F_{1,11}$= 7.335<br>P= 0.020 | $k_1$= -4.377<br>$b_1$= 6.603<br>$k_2$= 5.179<br>$b_2$= -7.405<br>$F_{1,11}$= 7.129<br>P= 0.022 | $k_1$= -4.304<br>$b_1$= 6.425<br>$k_2$= 5.082<br>$b_2$= -7.202<br>$F_{1,11}$= 6.642<br>P= 0.026 |
| Hybrid path/IC-based similarity Formula | Zhou | $k_1$= -0.817<br>$b_1$= 1.538<br>$k_2$= 1.100<br>$b_2$= -1.820<br>$F_{1,11}$= 8.266<br>P= 0.015 | $k_1$= -0.872<br>$b_1$= 1.428<br>$k_2$= 1.133<br>$b_2$= -1.689<br>$F_{1,11}$= 7.308<br>P= 0.021 | $k_1$= -0.851<br>$b_1$= 1.345<br>$k_2$= 1.081<br>$b_2$= -1.575<br>$F_{1,11}$= 7.025<br>P= 0.023 | $k_1$= -0.596<br>$b_1$= 1.011<br>$k_2$= 0.798<br>$b_2$= -1.214<br>$F_{1,11}$= 10.345<br>P= 0.008 | $k_1$= -0.376<br>$b_1$= 0.663<br>$k_2$= 0.808<br>$b_2$= -1.094<br>$F_{1,11}$= 4.280<br>P= 0.063 | $k_1$= -0.531<br>$b_1$= 0.888<br>$k_2$= 0.560<br>$b_2$= -0.916<br>$F_{1,11}$= 7.037<br>P= 0.022 | $k_1$= -0.524<br>$b_1$= 0.846<br>$k_2$= 0.695<br>$b_2$= -1.017<br>$F_{1,11}$= 6.077<br>P= 0.031 |
| Path-based similarity Formula |  | Wu–Palmer | Li | Al-Mubaid–Nguyen | Leacock–Chodorow | Rada |  |  |
| Path-based similarity Formula |  | $k_1$= -2.415<br>$b_1$= 3.533<br>$k_2$= 2.929<br>$b_2$= -4.046<br>$F_{1,11}$= 5.763<br>P= 0.035 | $k_1$= -3.767<br>$b_1$= 4.912<br>$k_2$= 3.669<br>$b_2$= -4.813<br>$F_{1,11}$= 5.089<br>P= 0.045 | $k_1$= -2.028<br>$b_1$= 2.614<br>$k_2$= 1.815<br>$b_2$= -2.400<br>$F_{1,11}$= 4.767<br>P= 0.052 | $k_1$= -1.720<br>$b_1$= 2.224<br>$k_2$= 1.528<br>$b_2$= -2.031<br>$F_{1,11}$= 4.555<br>P= 0.056 | $k_1$= -0.712<br>$b_1$= 0.997<br>$k_2$= 0.670<br>$b_2$= -0.954<br>$F_{1,11}$= 4.423<br>P= 0.059 |  |  |